\documentclass[twocolumn]{svjour3}
\usepackage[utf8]{inputenc}
\usepackage{hyperref}
\usepackage{times}
\usepackage{latexsym}
\usepackage{multirow}
\usepackage{color}
\usepackage{xcolor}
\usepackage{float}
\usepackage{url}
\usepackage{array}
\usepackage{latexsym}
\usepackage{amsmath}
\usepackage{mathtools}
\usepackage{amsfonts}
\usepackage{graphicx}
\usepackage{caption}
\usepackage[sort, numbers]{natbib}

\setcitestyle{square}

\begin{document}

\title{Lightme: Analysing Language in Internet Support Groups for Mental Health}

\author{Gabriela Ferraro \and Brendan Loo Gee \and Shenjia Ji \and Luis Salvador-Carulla}


\institute{Gabriela Ferraro \at Commonwealth Scientific and Industrial Research Organization \& Australian National University \\
           GPO Box 1700 CANBERRA ACT 2601 \\
           \email{gabriela.ferraro@data61.csiro.au} \\
           \and Brendan Loo Gee
           \at Australasian Institute of Digital Health \& Research School of Population Health, Centre for Mental Health Research, Australian National University \\
           \email{brendan.loogee@anu.edu.au}
           \and Shenjia Ji
           \at College of Engineering and Computer Science, Australian National University \\
           \email shenjia.ji@anu.edu.au
           \and Luis Salvador-Carulla
           \at Research School of Population Health, Centre for Mental Health Research, Australian National University \\
           \email{luis.salvador-carulla@anu.edu.au}
}

\date{Received: date / Accepted: date}

\maketitle

\begin{abstract}

\textbf{Background}: Assisting moderators to triage harmful posts in Internet Support Groups is relevant to ensure its safe use. Automated text classification methods 
                     analysing the language expressed in posts of online forums is a promising solution. \\
\textbf{Methods}:    Natural Language Processing and Machine Learning technologies were used to build a triage post classifier using a dataset from {\tt{Reachout.com}}
                     mental health forum for young people. \\
\textbf{Results}:    When comparing with the state-of-the-art, a solution mainly based on features from lexical resources, received the best classification performance 
                     for the \textit{crisis} posts (52\%), which is the most severe class. Six salient linguistic characteristics were found when analysing the crisis post; 
                     1) posts expressing hopelessness, 2) short posts expressing concise negative emotional responses, 3) long posts expressing variations of emotions, 
                     4) posts expressing dissatisfaction with available health services, 5) posts utilising storytelling, and 6) posts expressing users seeking advice 
                     from peers during a crisis. \\
\textbf{Conclusion}: It is possible to build a competitive triage classifier using features derived \textit{only} from the textual content of the post. Further research needs 
                     to be done in order to translate our quantitative and qualitative findings into features, as it may improve overall performance. \\

\end{abstract}

\section{Introduction}

Internet Support Groups (ISG) has been important and popular technologies for individuals with mental ill-health to receive support from other peers that have similar lived experiences 
\cite{Islam2018} and to anonymously share their stories with others to support their recovery \cite{mikal-hurst-conway:2017:CLPsych}. They are also referred to as online peer-support 
forum or networks. ISGs have supported groups of people with specific chronic health conditions, such as diabetes or mental health \cite{Islam2018, Naslund2018}. Current evidence suggests 
ISGs may have a positive impact on individuals with mental ill-health; however, it may also exacerbate a person’s distress levels \cite{Kaplan:2011}. Nevertheless, the safe use of ISGs 
will require more attention, especially designing mechanisms that can assist in mitigating possible adverse effects and harm to ISG users \cite{Griffiths:2017}.

Assessment and monitoring in ISGs are challenging and costly because it relies on the manual detection of posts in an online forum by trained moderators. This raises particular concerns 
on the scalability of ISGs as a potential digital health intervention. To overcome limitations, Natural Language Processing (NLP) and Machine Learning (ML) technologies can be used to 
build systems that can assist trained moderators in detecting and responding to hazardous posts that may cause further distress or self-harm to ISG users. Prior research has shown text 
classification methods to be promising solutions in reducing the workload of trained moderators \cite{Huh:2013}.

Moderators play an important role in managing communication between users in an ISG. They offer a range of informal support and advice to users, including providing personal experiences 
of recovery, motivating users to participate in the discussion, and enhancing the adoption of digital mental health services \cite{Kornfield:2018}. However, moderators may lack the 
necessary skills and expertise to guide appropriate decision making on issues relating to clinical safety \cite{Hartzler:2011}. Triaging ISG posts to assist moderators in reviewing 
new content uploaded daily is an automated text classification task designed to efficiently detect individuals’ thoughts, feelings, emotions, and possible behaviors represented in 
messages \cite{Conway:2016,Tausczik:2010}. 

Previous research has often focused on evaluating the performance of different ML classification models using mental health ISG data, such as Logistic Regression \cite{Cohan:2016, 
Pink:2016, Zirikly:2016}, Stochastic Gradient Descent (SGD) \cite{kim-EtAl:2016:CLPsych}, and Linear Discriminant Analysis (LDA) \cite{Shickel:2016}. The study by  \cite{Islam2018} 
used different ML techniques to detect depression from Facebook data. They evaluated the performance of several classification models, including Support Vector Machines (SVM), Decision 
Tree (DT), ensemble methods, and K-Nearest Neighbor (KNN). The results demonstrated the relative performance for specific classifiers. However, the authors did not evaluate the 
performance of different language features from lexical resources and deep learning models. 

Lexicon-based resources are central to modelling the linguistic characteristics of ISGs. Over the years, examples of comparative systems have used different lexicon-based features to 
classify hazardous posts in an ISG for mental health \cite{CLPsych:2017, milne-EtAl:2016:CLPsych}, including in posts from Twitter data \cite{Odea:2017, Odea:2015, 
coppersmith-EtAl:2016:CLPsych, jamil-EtAl:2017:CLPsych}. While modelling linguistic characteristics are important for accuracy performance, other features such as interactions of ISG users, 
forum structure, meta-data and other external features may likely improve the prediction performance \cite{Carron-Arthur:2015, Smithson:2011}. However, some authors have stated that 
relying on features extracted from external sources (e.g., forum structure and meta-data) may introduce biases; therefore, decreasing the predictive capabilities of the classifier on 
never seen before messages published on online forums \cite{Altszyler:2018}.

Our study focuses on developing an automated classifier for triaging posts using only features from the \textit{textual content} of the post derived from lexicon-based resources. We want to 
investigate the language of ISGs that exclude the use of the forum structure or post threads. By excluding forum structure and meta-data features from the model, the study primarily 
focuses on optimizing the linguistic aspects of detecting forum posts to avoid biases on unseen messages. Furthermore, given the extent of previous research on the combination of 
different ML classification models, we want to experiment on a broad combination of features using only a relatively small number of linear and nonlinear ML techniques including a 
couple of different deep learning models. 

\subsection{Research rationale}

This study used state-of-the-art methods to develop hand-crafted features derived from the {\tt{Reachout}} online support forum \cite{CLPsych:2017}. The model aims to achieve the best 
classification performance for \textit{crisis} posts and competitive results for other classification labels, described in Section \ref{materials}. We conducted a qualitative analysis of the post, 
which requires the immediate attention of moderators. The study has two aims:

\begin{itemize}

\item Shed some light about the linguistic characteristics of the urgent posts.
\item Examine the feasibility of lexical resources in an ML classification system for triage post using the {\tt{Reachout}} dataset. 

\end{itemize}

\section{Materials and Methods}
\label{materials}

\subsection{Dataset}
\label{datasetsection}

This study used a collection of posts from the Australian Reachout mental health online forum released by the Computational Linguistics and Clinical Psychology Shared Task 
(CLPsych) \cite{CLPsych:2017}. Participants range from 18 to 25 years old. All of the posts are written in English. Each post in the dataset is labelled with a semaphore 
pattern to indicate the urgency of the post, and the required attention of the moderator, as shown in Table \ref{table:labels}. Label distribution across the training and 
testing dataset of the Reachout online forum is given in Table \ref{table:dataset}.

\begin{table*}[ht]
\centering
\begin{tabular}{c p{6cm} p{6cm}}
Label & Description & Example \\ \hline
\textit{Green}  & No input from a moderator, and it can be safely left for the wider community of peers to respond. & \textit{I'm proud that I was able to call and keep up a 
                  phone conversation with my mum}. \\ \hline
\textit{Amber}  & A moderator should address the post at some point, but they do not need to do so immediately. & \textit{There are so many stuff I’m thinking about, but my 
                  medications are slowing my thoughts down and making it more manageable}. \\ \hline
\textit{Red}    & A moderator should respond to the post as soon as possible. & \textit{I feel helpless and things seem pointless. I hate feeling so down}. \\ \hline
\textit{Crisis} & The author, or someone they know, is in imminent risk of being harmed, or harming themselves or others. Posts should be prioritized above all others. & 
                  \textit{Im having some strong thoughts about ending my life, nothing helps}. \\ \hline
\end{tabular}
\caption{Severity label descriptions and examples in the {\tt{Reachout}} dataset.}
\label{table:labels}
\end{table*}

\begin{table}
\centering
\begin{tabular}{c c c c c} \hline
       & Train & \%                          & Test & \%                         \\ \hline
Crisis & 40    & \textcolor{darkgray}{3.36}  & 42   & \textcolor{darkgray}{10.5} \\   
Red    & 137   & \textcolor{darkgray}{11.53} & 48   & \textcolor{darkgray}{12}   \\
Amber  & 296   & \textcolor{darkgray}{24.91} & 94   & \textcolor{darkgray}{23.5} \\
Green  & 715   & \textcolor{darkgray}{60.18} & 216  & \textcolor{darkgray}{54}   \\ \hline
Total  & 1188  & -                           & 400  & -                          \\ \hline
\end{tabular}
\caption{Label distribution across training and testing set of the {\tt{Reachout}} dataset 2017}
\label{table:dataset}
\end{table}

Precision, recall and F-measure were used to examine the performance of the classifier. Precision is defined as the proportion of correctly classified posts into a particular label 
by the ML model. Recall is defined as the proportion of the labels that are successfully classified. The F-measure is the mean of precision and recall. Macro f-score metric is preferred 
since it gives more weight to infrequent yet more critical labels, such as \textit{red} and \textit{crisis}. Similar to \cite{Altszyler:2018}, the f-score for \textit{crisis}  
versus \textit{non-crisis} was reported. This metric measured the classifier’s capability to detect the most severe cases. Details of the official evaluation matrices are described below; 

\begin{itemize}
\item \textit{Macro-averaged F-score}: The macro-averaged f-score is calculated among \textit{crisis}, \textit{red} and \textit{amber}, and after excluding the \textit{green} class. 
\item \textit{F-score for flagged vs. non-flagged}: This metric separates the posts that moderators need to action (i.e. \textit{crisis}, \textit{red}, \textit{amber}) compared to 
              posts that can be safely ignored (i.e. \textit{green}). This is the most important metric in CLPsych since it measures the classifier’s capability to identify the post 
              that requires moderator attention.
\item \textit{F-score for urgent vs. non-urgent}: This metric is the average F1-score among urgent (\textit{crisis} + \textit{red}) and non-urgent (\textit{amber} + \textit{green}) labels.
\end{itemize}

A search of key computing and health databases (IEEE, ACM, PubMed and PsycINFO) were conducted to identify the key components of previous text classifiers for ISGs. Table 
\ref{table:rel-work} shows the features and methods used by the best performing classifiers using the Reachout dataset, more details in \cite{milne-EtAl:2016:CLPsych} and 
\cite{CLPsych:2017}. 

\begin{table*}
\centering
\begin{small}
\begin{tabular}{p{7cm} p{2cm}}
\hline
\textbf{Lexicon Features}                                 & \textbf{Used by}                       \\ \hline
        LIWC lexicon \cite{Pennebaker:2015}               & \cite{Cohan:2016,Malmasi:2016}         \\
        MPQA lexicon \cite{PHQ-9:Kroenke2001}             & \cite{Cohan:2016,Altszyler:2018}       \\
        PERMA lexicon \cite{Perma:2016}                   & \cite{Altszyler:2018}                  \\
        Emolex lexicon \cite{Mohammad2013CrowdsourcingAW} & \cite{Altszyler:2018}                  \\
        DepecheMood lexicon \cite{staiano2014depeche}     & \cite{Cohan:2016,Altszyler:2018}       \\ \hline
\textbf{Other Features}                                   &                                        \\ \hline
        Lexical diversity                                 & \cite{Altszyler:2018}                  \\
        Topic modeling                                    & \cite{Cohan:2016}                      \\
        TF-IDF weighted                                   & \cite{kim-EtAl:2016:CLPsych,Brew:2016} \\ 
        Character embeddings                              & \cite{Malmasi:2016}                    \\
        Word embeddings                                   & \cite{kim-EtAl:2016:CLPsych, 
                                                                  Brew:2016, Malmasi:2016, 
                                                                  Altszyler:2018}                  \\
        Sentence embeddings                               & \cite{Le:2014}                         \\
        POS-tags                                          & \cite{Malmasi:2016}                    \\ 
        Pronouns                                          & \cite{Altszyler:2018}                  \\
        Sentiment analysis                                & \cite{Shickel:2016,Zirikly:2016}       \\
        Post author                                       & \cite{Altszyler:2018}                  \\
        Post history                                      & \cite{Malmasi:2016,Altszyler:2018}     \\
        Post reply chain                                  & \cite{Pink:2016}                       \\
        Time of the post                                  & \cite{Altszyler:2018}                  \\
        Time between post                                 & \cite{Altszyler:2018}                  \\
        Week day of the post                              & \cite{Altszyler:2018}                  \\
        References to advisors                            & \cite{Altszyler:2018}                  \\
        References to self-harm                           & \cite{Altszyler:2018}                  \\
        References to Telephone helplines                 & \cite{Altszyler:2018}                  \\ \hline
\textbf{Algorithm}                                        &                                        \\ \hline 
        LDA: unsupervised topic modeling                  & \cite{Shickel:2016}                    \\
        SGD: supervised classification                    & \cite{kim-EtAl:2016:CLPsych}           \\
        Support Vector Machine (SVM): supervised classification & \cite{Malmasi:2016, Brew:2016, 
                                                                  Zirikly:2016, Altszyler:2018}    \\
        Logistic regression: supervised classification    & \cite{Cohan:2016, Pink:2016, 
                                                                  Zirikly:2016}                    \\ \hline
\end{tabular}
\caption{Examples of features used for triage classification using the {\tt{Reachout}} dataset. \textit{Used by} refers to previous research studies that a feature was used.}
\label{table:rel-work}
\end{small}
\end{table*}

\subsection{Predicting Alerts Approach}
\label{predicting_alerts}

The {\tt{Reachout}} dataset $D = \{x_i, y_i\}^n_{i=1} $ consist of $n$ training instances, where the $i$th instance is a feature vector $x_i$ and label $y_i$. The classification 
task is to predict the label $y_i$ given the feature vector $x_i$ for each forum post such that:

\begin{equation}
\hat{y_i} = \mathop{\arg\max}_{y_i}P_{\theta}(y_i | x_i)
\end{equation}

We trained a Support Vector Machine (SVM) multi-class classifier with linear kernels \cite{Vapnik:1963}. SVM is a supervised ML method used widely in text classification. This method used a 
state-of-the-art triage classification using the {\tt{Reachout}} dataset. Hyper-parameters\footnote{In machine learning, an hyper-parameter is a parameter whose value is set before the 
learning process, while the value of other parameters are derived via learning.} were selected with a grid search\footnote{Grid search is a way of choosing the best 
hyper-parameters, and consist of exhaustively searching through a subset of the hyper-parameter space of a learning algorithm.} scheme with a 5-fold Cross-Validation over the training set. 
The C hyper-parameter\footnote{The C hyper-parameter referrers to the regularization value, which serves as a degree of importance that is given to miss-classification. 
The larger the value, the less the wrongly classified examples are allowed.} is 1 with $l_1$ regularization type, and the loss function\footnote{A loss function or cost function measures 
how good a prediction model does in terms of being able to predict the expected outcome.} is \textit{hinge}, the maximum number of iterations is 2000.

In order to compare Lightme (which used SVM) against other ML classifiers, we also trained K-Nearest Neighbour (KNN), and Naïve Bayes. Since deep learning models are the state-of-the-art 
in many natural language processing applications, we trained two neural network classifiers: Multi Layer Perceptron (MLP) and Recurrent Neural Networks (RNN) with Long Short Term Memory (LSTM).

The feature set is shown in Table \ref{table:features}. All the features were derived from the post themselves. No features derived from the forum structure or interactions between posts 
were used. We included additional language features such as MPQA, offensive language, and mental health lexicons.

During feature extraction, negation was model as in \cite{cimino2014linguistically}. Thus, when a $term_i$ from a post is found in a lexicon, its negation is checked by inspecting the 
term $term_{i-1}$. As in \cite{cimino2014linguistically}, we used a list of negation terms:

\begin{minipage}{8cm}
\textit{no}, \textit{nobody}, \textit{nothing}, \textit{none}, \textit{never}, \textit{neither}, \textit{nor}, \textit{nowhere}, \textit{hardly}, \textit{scarcely}, \textit{barely}, 
\textit{don't}, \textit{isn't}, \textit{wasn't}, \textit{doesn't}, \textit{ain't}, \textit{can't}, \textit{won't}, \textit{wouldn't}, \textit{shouldn't}, \textit{couldn't}, 
\textit{hasn't}, \textit{haven't}, \textit{didn't}
\end{minipage}

If a negation term was found, the polarity of the term was shifted when the lexicon differentiate between positive and negative terms (e.g., the PERMA lexicon); otherwise, it was skipped 
and not included as a feature.

\begin{table*}
\centering
\begin{small}
\begin{tabular}{p{4cm} p{9cm}}
\hline 
\textbf{Lexicon Features}                                                                                         &  \textbf{Feature Description}                                                                                   \\ \hline \hline
        MPQA lexicon*                                                                                             & The number of words with MPQA polarity in each post                                                             \\ \hline
        DepecheMood lexicon*                                                                                      & The number of words overlap between each category in DepechMood and a post.                                     \\ \hline 
        Emolex lexicon*                                                                                           & The number of words overlap between each category in the NRC-Emotion-Lexicon-v0.92 lexicon and a post           \\ \hline
        Mental Disorder lexicon
        \footnote{\url{http://mental-health-matters.com/psychological-disorders/alphabetical-list-of-disorders/}} & The number of words overlap between the Mental Disorder lexicon and a post                                      \\ \hline 
        PHQ\_9 lexicon                                                                                            & The number of words overlap between the PHQ\_9 and a post                                                       \\ \hline
        PERMA lexicon (1)*                                                                                        & The number of bi-gram and tri-gram overlap between PERMA and a post                                             \\ \hline
        PERMA lexicon (2)                                                                                         & The number of bi-gram and tri-gram overlap between PERMA negatives categories and a post                        \\ \hline
        PERMA lexicon (3)                                                                                         & The weights sum of the bi-gram and tri-grams overlap between PERMA and a post                                   \\ \hline
        Offensive word lexicon\footnote{\url{https://www.cs.cmu.edu/~biglou/resources/}}                          & The number of words overlap between offensive word list                                                         \\ \hline \hline
\textbf{Other Features}                                                                                           &                                                                                                                 \\ \hline \hline
        TF-IDF weighted                                                                                           & N-grams TF-IDF representation of each post with top max features chosen by Scikit-learn based on term frequency \\ \hline
        Pronouns                                                                                                  & The number of pronouns used in each post, including \textit{I}, \textit{me}, \textit{you}, \textit{he}, 
                                                                                                                    \textit{him}, \textit{she}, \textit{her}, \textit{it}, \textit{we}, \textit{us}, \textit{they}, \textit{them}   \\ \hline
        Mean word length                                                                                          & The average length of words in a post                                                                           \\ \hline
        Sentence embeddings                                                                                       & Sentence representation computed by averaging pre-trained FastText word embeddings fine-tuned with 
                                                                                                                    the Reachout dataset                                                                                            \\ \hline
        Last sentence embeddings                                                                                  & Sentence representation of the last sentence in each post computed by averaging FastText word embeddings 
                                                                                                                    trained with the Reachout dataset                                                                               \\ \hline
        Sentiment analysis feature                                                                                & The sentiment of each post classified by a sentiment classifier trained by us with GloVe 
                                                                                                                    \citep{pennington2014glove} word embeddings feature and emoticon embedding                                      \\ \hline
        User rank                                                                                                 & The forum title of the poster for each post                                                                     \\ \hline
        Number of web links                                                                                       & Total number of web links in a post                                                                             \\ \hline
        Number of reference to a help line services                                                               & \textit{mental health}, \textit{australia}, \textit{general practitioner}, \textit{doctor}, 
                                                                                                                    \textit{psychologist}, \textit{counsellor}, \textit{gp} (general practitioner), \textit{emergency}, 
                                                                                                                    \textit{000}, \textit{lifeline}, 131114, 13 11 14, \textit{kids help line}, 1800 55 1800, 1800551800, 
                                                                                                                    \textit{salvation army care line}, 1300 36 36 22, 1300363622, \textit{e-couch}, \textit{moodgym}, 
                                                                                                                    \textit{bluepages}, \textit{black dog institute}, \textit{reachout}, \textit{beyondblue}, 
                                                                                                                    \url{www.moodgym.anu.edu.au}, \url{www.ecouch.anu.edu.au},  \url{www.bluepages.anu.edu.au}, 
                                                                                                                    \url{www.researchout.org.au}, \url{www.blackdoginstitute.org.au}                                                \\ \hline
        Number of references to self-harm expressions                                                             & \textit{suicide}, \textit{kill myself}, \textit{kill my self}, \textit{cut myself}, \textit{cut my self}, 
                                                                                                                    \textit{hurt myself}, \textit{hurt my self}, \textit{harm myself}, \textit{harm my self}, 
                                                                                                                    \textit{I want to die}, \textit{don't want to live}, \textit{end my life}, \textit{kill}, \textit{hurt}, 
                                                                                                                    \textit{cut}, \textit{want to die}, \textit{I don't want to live}                                               \\ \hline
        Number of references to advisors                                                                          & \textit{supervisor}, \textit{supervisors}, \textit{mentor}, \textit{manager}, \textit{tutor}, 
                                                                                                                    \textit{case-manager}, \textit{managers}, \textit{manager}, \textit{psych}, \textit{psychiatrist}, 
                                                                                                                    \textit{gp} (general practitioner), \textit{gps}, \textit{counsellor}, \textit{counselor}                       \\ \hline \hline
\end{tabular}
\caption{Feature set used for triage classification with the {\tt{Reachout}} dataset. '*' indicates a lexicon that have been tested in the previous studies (see Table~\ref{table:rel-work}).}  
\label{table:features}
\end{small}
\end{table*}

\section{Results}
\label{results}

\subsection{Triage Classification Experimental Results}
\label{experiment}

The results of the triage classification experiment using the different features, including lexical resources and treating negation, are presented in Table \ref{table:features-sets-results}. 
Best results are highlighted in boldface. Exclusive use of lexicon features resulted in lower performance for all classes (\textit{flagged}, \textit{urgent} and \textit{crisis}) and 
the overall performance (\textit{macro F1-score}). Treating negation when using only lexicons did not boost the classification performance. However, adding Term Frequency-Inverse 
Document Frequency (TF-IDF) contributed to improving the classification performance for all classes\footnote{TF-IDF is the amount of times a word appears in a document weighted by the 
number of meaningful words across multiple documents}. Best results were achieved with features that included “TF-IDF + lexicons with negation”, F1-score of 0.44. 
The most complex set of features (included all features in Table \ref{table:features}) showed competitive results with the most state-of-the-art triage classification system 
by \cite{Altszyler:2018}, and the baseline classification system by \cite{milne-EtAl:2016:CLPsych}. \newline

\begin{table*}
\begin{small}
\centering
\begin{tabular}{lllll}
\hline
                                                    & Macro F1-score & Flagged       & Urgent         & Crisis        \\ \hline
Only lexicons                                       & 0.24           & 0.38          & 0.38           & 0.20          \\
Lexicons with negation                              & 0.19           & 0.43          & 0.37           & 0.04          \\
TF-IDF + lexicons                                   & 0.38           & 0.71          & 0.53           & 0.44          \\ 
TF-IDF + lexicons with negation                     & \textbf{0.44}  & 0.74          & \textbf{0.63}  & \textbf{0.52} \\ 
Lightme (features from Table \ref{table:features})  & 0.43           & \textbf{0.77} & 0.59           & \textbf{0.51} \\ \hline
\end{tabular}
\caption{Triage classification with different features sets.}
\label{table:features-sets-results}
\end{small}
\end{table*} 

We also experimented on a linear and nonlinear classification method. Naïve Bayes was trained with the Lightme feature set since it is an easy and fast linear classification 
method suitable for classifying large chunks of data. Similarly, KNN was trained with the same feature set due to its practicality and ease. Hyper-parameters such as the number 
of neighbours were selected with a grid search scheme using a range from 1 to 25. Table \ref{table:official_results} shows the results of Naïve Bayes and KNN, which underperformed 
SVM and other state-of-the-art systems. Compare to Naïve Bayes and KNN, SVM is known to perform better on rich feature sets such as the one presented in this study. 

MLP was trained using the same set of features as Lightme with hidden layer sizes that varied between 100 and 300 nodes, depending on the development set. The RNN+LSTM model was 
trained with pre-trained word embeddings and without features since one of the advantages of this type of model is its ability to learn feature representations automatically. 
Important hyper-parameters such as the number of epochs, size of the hidden layer and batch size were tuned using a portion of the training set as the development set. 
As shown in Table \ref{table:official_results}, the deep learning models underperformed the other models. This is not surprising as deep learning models are data-hungry 
and the size of Reachout is small, especially some of the classes (e.g., red and crisis) only have a few instances.

\begin{table*}
\begin{small}
\centering
\begin{tabular}{l l l l l}
\hline
\textbf{System}                 & \textbf{Macro F1-score} & \textbf{Flagged} & \textbf{Urgent} & \textbf{Crisis} \\ \hline
Baseline                        & 0.3                     & 0.61             & 0.44            & -               \\
Naïve Bayes                     & 0.28                    & 0.67             & 0.42            & 0.39            \\
KNN                             & 0.14                    & 0.39             & 0.08            & 0.0             \\
MLP                             & 0.38                    & 0.71             & 0.58            & 0.39            \\
RNN+LSTM                        & 0.28                    & 0.44             & 0.008           & 0.0             \\
Altszyler \cite{Altszyler:2018} & 0.44                    & 0.90             & 0.68            & 0.48            \\
TF-IDF + lexicons with negation & 0.44                    & 0.74             & 0.63            & \textbf{0.52}   \\ 
Lightme (features from Table 
\ref{table:features})           & 0.43                    & 0.77             & 0.59            & \textbf{0.51}   \\ \hline
\end{tabular}
\caption{Comparison results on the test set in terms of F-score}
\label{table:official_results}
\end{small}
\end{table*}

\subsection{Qualitative formative analysis of \textit{crisis} posts}
\label{qualitative}

We randomly selected 40 crisis posts to analyse from the training dataset. We then used open coding to understand linguistic characteristics. Through the qualitative analysis of 
the selected crisis posts, we identified six linguistic characteristics. We extracted selected phases of crisis posts (including the post id) that matched the given linguistic profile, 
and suggested recommendation of features to the model. 

\subsubsection{Expressing hopelessness in crisis} 

Many of the posts used language or words that described a person’s feeling of immediate hopelessness. Extreme hopelessness or helplessness may be associated with an increased risk of 
suicide \cite{Cash2013}. Learned helplessness comes from a repeated belief that uncomfortable situations are inescapable, an example statement is ''\textit{I tried doing this for my anxiety, 
but I ended up faced with these challenges}'' \cite{Liu:2015}. Hopelessness is the feeling of a combination of helplessness and experiences of depression resulted from a person’s 
response to a negative event \cite{Cash2013, Liu:2015}. An example statement is ''\textit{I am fed up with my friend anger! I can’t bother trying anymore because I am frustrated}''. \newline 

\noindent Extracted phases of crisis posts describing hopelessness of a forum user:

\begin{itemize}
\item ''\textit{I can feel pretty hopeless at times too. I start questioning if I can ever get better. It's hard enough to live.}'' (Post ID: 136600)
\item ''\textit{I'm feeling so tired, and I want to give up on life. I need to keep holding on. There's still hope for me. I just need to make sure I reach out when I feel like things 
                are getting way too intense.}'' (Post ID: 136601)
\item ''\textit{No but I am pretty friggin sick of my entire life at this point and my existence...}'' (Post ID: 135818)
\item ''\textit{I'm still finding it hard not to do anything stupid. I've screwed up. Now I don't know where this is headed.}'' (Post ID: 138188)
\end{itemize}

\noindent\textbf{Recommended features to model}: Categorical features can be model with the following keywords; 
\textit{feel tired, fed up, better dead, give up life, the end is near, sick of life, sick of existence, holding on, hopeless times, hope, trying help or talking,} 
and \textit{hard to try or do}. Other features can include checking spelling mistakes.

\subsubsection{Short crisis posts and emotional response}

Short length posts contained concise descriptions of a person’s negative emotions. Contrast to longer post; shorter posts contained more variations in expressing positive and negative 
emotions. As noted by \cite{Odea:2017}, lexicons may be limited in detecting certain expressions such as irony, sarcasm, and metaphors. Therefore, any text under 50 words should be 
interpreted with caution. Further limitations of interpreting short posts included the use of negation \cite{gkotsis-EtAl:2016:CLPsych2}. \newline

\noindent Examples of short crisis posts describing a concise negative emotion:

\begin{itemize}
\item ''\textit{I'm suffocating. I don't if I can do this anymore.}'' (Post ID: 138064)
\item ''\textit{@redhead I don't know how long I can even keep myself together before I'm screwed.}'' (Post ID: 138067)
\item ''\textit{@chessca\_h no. I don't want to be safe anymore. I'm ai over it right now.}'' (Post ID: 137786)
\end{itemize}

\noindent\textbf{Recommended features to model}:  Features can define short posts as messages under 50 words, or posts that contain no more than two sentences. Additionally, 
features can include detecting only one negative emotion for short posts, treating all negations for short posts, and checking for spelling mistakes.

\subsubsection{Long crisis posts and emotional coping}

Long length posts were found to start with a user expressing some negative emotions, followed by positive emotions related to their abilities to cope. This may be a positive sign 
as it may indicate a person attempting to reconcile negative emotions \cite{Naslund2014}. However, they may express negative emotions after showing signs of positivity. \newline

\noindent Extracted phases of crisis posts that expresses patterns of health service dissatisfaction:

\begin{itemize}
\item ''\textit{\underline{(Negative)} Feeling extremeley tired each morning. It's getting to the point I'm contemplating ringing in sick to aviod getting up. 
        \underline{(Positive)} \textbf{Despite the tiredness, I've bee getting up and going to work, because I know I need to face the world.} 
        \underline{(Negative)} Keep having thoughts to end it all...}'' (Post ID: 135898)
\item ''\textit{\underline{(Negative)} Didn’t sleep til way after 2 last night. It was super hard to sleep, then wake this morning. I just wanted to ignore the world today. 
        \underline{(Positive)} \textbf{I eventually fell asleep. I eventually got up and went to work. I faced the world and smiled a little.} 
        \underline{(Negative)} Really struggled through my shift today... }'' (Post ID: 137919)
\end{itemize}

\noindent\textbf{Recommended features to model}: Feature can define long posts as messages that contain 50 or more words with varying levels of positive and negative emotions. 
Other features can include detecting negative emotions at the beginning of a sentence followed by subsequent positive emotions. Checking for spelling mistakes can also be a feature. 
A feature can detect positive emotions such as keywords relating to coping; \textit{getting there, I faced the world,} or \textit{getting up and working}.

\subsubsection{Health service dissatisfaction} 

It was found that people who seek support services (e.g., health service, counsellors, or treatment) in the forum would sometimes feel hopelessness, avoidance, or frustration. 
This pattern may signal a person using the forum to vent their dissatisfaction or frustrations with local mental health support services, or failures with their 
treatment of care. \newline

\noindent Extracted phases of crisis posts that expresses patterns of health service dissatisfaction: 

\begin{itemize}
\item ''\textit{\underline{(Service)} \textbf{My gp} was running late today, which heightened my anxiety. At first she didn’t realise it was my 3month follow up apt. She also had no 
        idea that the psych was meant to write a letter, so the psych either ran out of time or forgot. Meh. I just wanted to run away and hide. It was SO VERY hard not to close off 
        and run. I didn’t even hear her call my name the first time. Blergh...}'' (Post ID: 137919)
\item ''\textit{im having bad thought about ending my life, nothing helps not even \underline{(Service)} \textbf{my counceller}}'' (Post ID: 136895)
\item ''\textit{I know looking back at therapy experiences that didn't work out will only discourage me. I'm highly impatient and annoyed. I'm trying to find the right 
        \underline{(Service)} \textbf{professionals for me}, its a very frustrating process. I can feel pretty hopeless at times too. I start questioning if I can ever get better. 
        It's hard enough to live.}'' (Post ID: 136600)
\end{itemize}

\noindent\textbf{Recommended features to model}: Categorical features can be used to model any mention of support services. Features can also consider detecting the negative 
expression of support services in crisis posts, factoring the length of the posts, and checking for spelling mistakes.

\subsubsection{Utilising story telling to express crisis} 

According to \cite{Smithson:2011}, there are two types of behaviours when ISG users seek help. The first behaviour involves a person wanting to communicate their story or 
’trouble telling’, and the other behaviour involves a person wanting advice. Appropriate timing for offering advice is crucial. If the advice is suggested too soon, it 
is likely to be rejected. It was found some people would join the forum to seek advice about some issue followed by opening up to talk about their problems. \newline

\noindent Extracted phases of story telling in crisis posts:

\begin{itemize}
\item ''\textit{...\underline{(Event)} \textbf{So today I went to the doctors and they told me that the chemotherapy that I am on is not working}, my body isnt reacting to 
        it the way it should be, which means that I now need to start this new treatment that is going to knock me around a lot more then the ast chemotheraphy...}'' 
        (Post ID: 136116)
\item ''\textit{\underline{(Event)} \textbf{I moved out of home into a defacto relationship about a year ago now}, and despite having troubles with my mum, who I used to live with 
        (single parent), I have the feeling that she is very lonely and she often gets teary about that. \underline{(Event)} \textbf{She mentioned today that she may as well just 
        kill herself because she feels like she's not really worth it anymore.}}'' (Post ID: 137384)
\end{itemize}

\noindent\textbf{Recommended features to model}: Features can detect a sequence of personal events. Personal events may contain temporal features, such as 
\textit{today, yesterday, or tomorrow}. Additionally, features can detect negative and positive emotional responses relating to different events identified in the post. 
Other features can include checking spelling mistakes. 

\subsubsection{Seeking advice of peers during crisis}

Crisis post was found to contain more advice seeking information than information providing support to other peers. Gaining the support of peers online is a common behaviour 
among people with severe mental ill-health \cite{Naslund2014}. \cite{Carron-Arthur:2015} differentiate posts that \textit{provide support to peers} and posts that attempt 
to \textit{seek advice from other peers} in an ISG. Supportive posts were characterized by the user providing emotional support, and informational support, such as offering 
website links to seek help. However, posts that seek advice are characterized by the users seeking informational support and seeking emotional support, and companionship 
from other peers. \newline

\noindent Extracted phases of advice seeking in crisis posts:

\begin{itemize}
\item ''\textit{...Suffering from anxiety and deppression myself, this kind of relationship is setting me back quite significantly. \underline{(Advice Seeking)} 
        \textbf{Has anyone else ever had a depressed parent that they are worried about when they move out of home?} I have been going to her place often and not sure what 
        else I can do to really help her...}'' (Post ID: 137384)
\item ''\textit{She's very depressed and always wnats to die. I'm pretty scared and I try to help but deep down I'm pretty useless for helping.\underline{(Advice Seeking)} 
        \textbf{Any good tips?} Because she likes to talk to me because i'm nice to her and doesn't judge her.}'' (Post ID: 135748)
\end{itemize}

\noindent\textbf{Recommended features to model}: Features can detect questions relating to an emotional response. Other feature can also detect advice-seeking information 
as a text embedding feature that identifies information relating to emotional support. 
\section{Discussion}
\label{discussion}

\subsection{Principal Findings}

This study demonstrates a solution that utilises a variety of lexicon-based resources and supervised ML techniques to assist trained moderators to efficiently moderate ISGs. 
Contrast to other similar research; this study extracted lexicon-based features from the textual content of posts which may avoid possible biases during classification. 
The classification experiment found one of our classifier (Liteme) achieved the best results for the \textit{crisis} post (0.52 F1-score) and competitive results in the 
other classes (i.e., \textit{non-green}, \textit{flagged}, and \textit{urgent} posts). These results may indicate that it is possible to build a strong classifier that 
can process only textual features extracted from individual messages. However, the experimental results also demonstrated that using only lexicons was not enough to 
classify posts into all relevant classes. Exclusive use of vocabulary in the {\tt{Reachout}} dataset was built into the solution which may have introduced some noise that may 
have impacted on the classification performance of \textit{flagged} and \textit{urgent} posts. Furthermore, this study demonstrated the limitations of utilizing lexicons, especially 
their ability to only capture information at the ’word’ level. This may prevent their ability to understand the contextual meaning at the ’sentence’ level. 

Furthermore, the findings suggest that using mental health lexicons can have an impact on the classification of posts requiring immediate response by trained moderators. This is 
unsurprising given the distinct domain-specific properties of lexicons, especially their association with certain mental and behavioural health theoretical constructs \cite{Kornfield:2018}. 
Lastly, six linguistic characteristics were identified in the qualitative analysis of \textit{crisis} posts. Interestingly, we found a person in crisis will use words or language 
associated to hopelessness, publish short posts containing concise negative emotional responses, publish long posts containing variations of emotions, express dissatisfaction 
with locally available health services, use storytelling to express crisis, and seek the advice of peers during a crisis.

\subsection{Comparison to Previous Research}

Our best classifier showed comparative results with the state-of-the-art systems for triage classification with the Reachout dataset and the baseline classifiers. The baseline system 
by \cite{milne-EtAl:2016:CLPsych} used uni-grams and bi-grams as features, and a default scikit-learn logistic regression classifier \cite{scikit-learn}. We also found the 
classification performance of the best system by \cite{Altszyler:2018} from the CLPsych 2017 Shared Task also utilized an SVM classifier. This classifier used a richer set of 
features, including features from the forum structure and interactions between posts, which outperformed our system for \textit{flagged} and \textit{urgent} posts. Interestingly, 
our approaches showed better results in identifying \textit{crisis} posts. This is an important category for this problem, especially the moderator’s need to immediately 
respond to these posts. Adding features derived from the forum structure may help to improve the classification performance. However, the trade-off is the expense of not 
properly classifying posts from new users.

Similar to other systems, our triage text classifier found using TF-IDF with lexicons improved classification performance. \cite{kim-EtAl:2016:CLPsych} received the best results 
in CLPsych 2016 Shared Task when using TF-IDF weighted n-grams and post embeddings using Sent2vec in an SGD classifier, and a set of twelve fine-coarse grained labels, instead of 
the coarse-grained four labels. The system by \cite{Brew:2016} also weights n-grams with TF-IDF producing similar results. Similarly, the use of TF- IDF showed comparative 
results to triage text classifiers for Twitter \cite{Odea:2015}, and an online support forum for substance abuse \cite{Kornfield:2018}.

The qualitative findings appeared to support prior research that found similar patterns of online interactions in people with mental ill-health using social media. As highlighted 
in previous research, online peer-to-peer interactions can improve health and psychosocial outcomes by facilitating a range of positive behaviours that can empower people, such as 
seeking information and emotional support \cite{Naslund2014}. However, these online networks can also become harmful when social media content begin to promote self-harm, suicide, 
or pro-eating disorder behaviours \cite{Gerrard2018, Dyson2016}. Particularly, social media posts that promote “problematic” content that may be difficult to identify by specifically 
moderating hashtags in online communities \cite{Gerrard2018}.

\subsection{Implication on Future Research}

Most of the qualitative findings may be translated into features that could improve classification performance. As noted, the qualitative findings of the \textit{crisis} posts could 
be used to distinguish salient linguistics characteristics of the language used in urgent messages for moderators. For example, specific features for detecting hopelessness may 
improve detection of crisis messages. Suggestions for future work may also include differentiating posts that provide support or seek advice to other peers and identifying 
participants roles, such as leaders, influence, and opinion users \cite{Carron-Arthur:2015}. Furthermore, various types of help-seeking behaviours can be identified, such as users 
wanting to share their personal stories of struggle \cite{Smithson:2011}. The analysis of satisfaction with available services can play a role in developing enhanced mixed 
reality care approaches combining eHealth and on-site services \cite{vanGenderen2018}. 

\subsection{Limitation}

There was a limitation to this study. First, our classifier is restricted to one dataset. More data is needed to generalize the model to avoid overfitting. Second, the training set 
was relatively small. This may have had implications to our approach and subsequent results. Third, an error analysis was not conducted. The error analysis could have examined why 
certain posts were misclassified or classified correctly. 

\subsection{Conclusion}
\label{conclusion}

The current study examines a triage classifier using features derived only from the textual content of the post. Various lexicons were used to analyse the value of lexical resources 
on the text classifier for triaging posts. Lexical resources alone were not enough to build a good performing classifier; however, a solution that includes lexicons with other features 
derived from the content of the posts performed well in identifying \textit{crisis} posts. Qualitative investigation on the \textit{crisis} posts found six salient linguistic characteristics. 
While qualitative findings are still formative, more work is needed to translate these findings into features that can improve the overall performance.

\section{Disclosure Statement}

No competing financial interests exist.



\bibliographystyle{spmpsci}      

\bibliography{biblio}

\end{document}